# Contribution of case based reasoning (CBR) in the exploitation of return of experience

## "Application to accident scenarii in rail transport"

Ahmed MAALEL[A] and Habib HADJ MABROUK[B]

A ISTLS
Higher Institute of Transport and Logistics of Sousse
maalel.ahmed@gmail.com
B INRETS
National Institute for Research on Transport and Safety
2 avenue du Général Malleret-Joinville, 94114 Arcueil Cedex - France Tél. : +33 612276612
mabrouk@inrets.fr

***Abstract*—The study is from a base of accident scenarii in rail transport (feedback) in order to develop a tool to share build and sustain knowledge and safety and secondly to exploit the knowledge stored to prevent the reproduction of accidents / incidents. This tool should ultimately lead to the proposal of prevention and protection measures to minimize the risk level of a new transport system and thus to improve safety. The approach to achieving this goal largely depends on the use of artificial intelligence techniques and rarely the use of a method of automatic learning in order to develop a feasibility model of a software tool based on case based reasoning (CBR) to exploit stored knowledge in order to create know-how that can help stimulate domain experts in the task of analysis, evaluation and certification of a new system.**

## I. INTRODUCTION

The importance of taking into consideration experience and lessons from the past are the basis for the establishment of a system of feedback (Rex) as one of the essential means to promote nature improving security. The experience feedback allows to learn from a lived experience in order to prevent its reproduction by implementing preventive measures to minimize the occurrence of situations of insecurity and adequate remedial measures to mitigate the severity of the damage caused. The Rex is thus a dynamic process of collection, storage, analysis and data exploitation for situations contrary to safety (accident or incident) [1]. However, until now, we do not know of a tool capable of helping field experts develop a new system or an analog system. The work that exist in these days to operate the Rex are mainly based on analysis of accident statistics and are quantitative and therefore do not take into account the semantic aspect of knowledge that represents an undeniable richness. However, if one wishes to respect the national and European regulations in force [2, 3,4,5,6], it seems essential to use the principle GAME (Overall At Least Equivalent) This principle requires that the development of a project must be at least equivalent in terms of security to the existing analog system that is known safe. The use of content of Rex may just be beneficial and provides a partial answer to the principle GAME advocated by the new regulations. Our study is within this context and aims to develop an expert system to assist the operating of experience feedback.

## II. THE RETURN OF EXPERIENCE

The first phase of the construction of the experience feedback focuses on the enumeration of all the anomalies encountered and the collection of maximum data. Data collection relate primarily to the human operator, to his internal and external environment namely the technical system [7]. The second phase, analysis, fulfills the principle of "understand" and helps identify the mechanisms generating events affecting safety. The storage and archiving of data collected comes after analysis in a database and often through a tool. The next phase is the operation that consists in using and interpreting the results from different information which main objective is to extract the truly predictive event , to consider isolated cases and to predict or imagine future scenarii of accidents or events not taken into account. Finally the phase of recommendations is to clarify and identify the measures of prevention and protection appropriate to limit the reproduction of an event of insecurity. It is better to take advantage of the lessons of experience to improve safety.

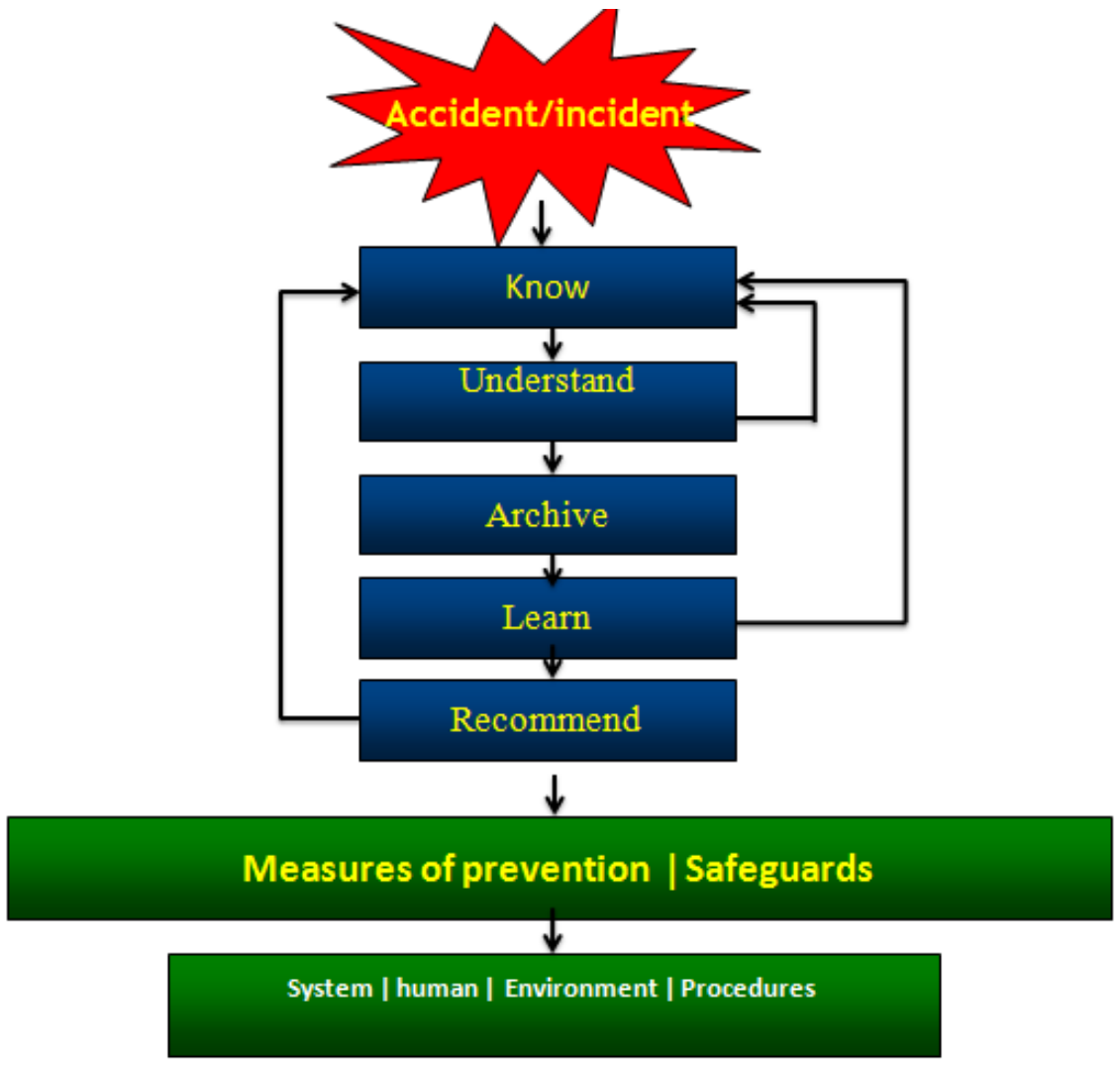

**Fig 1.** Articulation of the various stages of progress of Rex [1]

It is commonly accepted that feedback is of undeniable interest in reducing the level of risk and therefore in improving safety, and to accomplish this objective, data and knowledge from the Rex should be organized and well accounted for in the next phases of operation and recommendation. Research is increasingly experienced in process improvement REX, however, the majority of this work will explore the use of terminologies and formalisms with conventional tools. Little attention is given to the downstream stages of that particular operation. REX approach is usually limited to generating quantitative statistical reports on railway accidents. It is therefore for us to propose a way to exploit this knowledge in a qualitative way to prevent such scenarii from recurring in the future.
The approach to give a partial answer depends largely on the use of artificial intelligence techniques , including automatic learning systems.

## III. GENERALITIES ON LEARNING

### *A HUMAN LEARNING AND AUTOMATIC-LEARNING*

The ability to learn appears to be an essential component of the definition of intelligence which all human beings have the privilege of Indeed, man is pre-programmed to learn and he cannot do anything without memoring what he does, sometimes against his will. The newborn who has already attitudes and reflexes is fundamentally curious. Man, cannot help pursue information, all his life cannot help turne his eyes towards something that moves, cannot help to stretching his ear to an unusual sound [8]. The computer is a priori devoid of such a research and general learning program. In addition, methods and approaches to research currently developed through machine learning cannot, in any circumstances, replace the mechanisms of learning and intelligent behavior that men are endowed with indeed, most results obtained so far for machine learning are based on very simplistic assumptions compared to the actual mechanisms of human learning.

### *B. THE PURPOSE OF AUTOMATIC-LEARNING*

The emergence and development of industrial-based systems require knowledge of design tools to aid learning, including learning mechanisms. Learning indeed has created a growing interest in recent years, as evidenced proved by the impressive number of publications and conference which is the subject. This discipline, regarded as a promising solution to aid learning, including attempts to answer some questions [9]: how to represent an explicit body of knowledge, how to manage, enhanced, modify? According to GANASCIA [10], machine learning is defined by a double objective: an objective scientific : understanding and mechanizing the phenomena of evolution in time and the adaptability of reasoning and a practical objective : the automatically acquisition of knowledge bases from examples.

Indeed, there is so much different knowledge to get that the ideal should be an automatic system learns about himself from examples rather than to receive the human one by one. Learning can be defined by the performance improvement with experience. In fact, according to GANASCIA / 90 / [11], learning is intimately linked to the generalization: learning means passing a series of situations experienced in a reusable knowledge in similar situations.

In summary, learning is a very general term that describes the process by which the human or the machine can increase his knowledge. Learning is therefore reasoning, discovering analogies and similarities, generalization or particularization of experience, built on its past mistakes and failures for subsequent reasoning. The new findings are used to solve new problems, accomplish a new task or increase performance in the achievement of an existing task, to explain a situation or to predict behavior. The fields of human activity are increasingly complex and involve amounts of information that synthesize the human mind with difficulty. Extracting from this mass of data relevant and useful knowledge for explanatory purposes or decision is the main objective of machine learning. [12]

C. VARIOUS MODES OF REASONING IN LEARNING

The machine learning mechanism is based on four modes of reasoning or inference: induction, deduction, abduction and analogy. The purpose of this paragraph is to define these four terms in reference to definitions from work by learning and artificial intelligence. GRUNDSTEIN [13] defines these terms through an example:

**The deduction** proceeds of a rule and a fact to get results:
Rule: All beans from this bag are white
Fact: These beans are from this bag
Result: These beans are white.

**Induction** leads to a rule by starting with a fact and a result:

Fact: These beans are from this bag
Result: These beans are white
Rules: All beans from this bag are white

**Abduction** leads to the fact starting from the a rule and a result:

Rule: All beans from this bag are white
Result: These beans are white
Fact: These beans are from this bag.

- Deduction: From A and A => B is "offset" B
- Abduction: from B and A => B, on "abduit" A
- Induction: from A (z) => B and A (t) => B is "induced" A (x) => B

The analogy is used in practice to understand or interpret new situations from previous situations already stored. Analogy combines the notion of similarity (or resemblance) and the notion of causation (Figure 1). More formally, one has a situation analogous source of the form (A, B) and a target position of the form (A ', B'). There are relations of similarity (and dissimilarity) between A and A ', respectively B and B', and dependent relationships, generally causal nature between A and B, respectively A 'and B'.

D. *THE CASE BASED REASONING: CBR*

The Case Based Reasoning is a type of reasoning in AI in the field of automatic learning. Case based reasoning means remembering past situations similar to the current situation and by these situations to help resolve the current situation. The case based reasoning (CBR) is a form of reasoning by analogy.. [18] The analogy searches for cause and effect relation in past situations and transfer to the current situation the similarities to between then past situations as well as the current situation. The case based reasoning research only look for similarities or proximity relations between past situations and the current situation. The C.B.R. considers reasoning as a process of remembering a small set of practical situations: the cases, it bases its decisions on the comparison of the new situation (target cases) with the old (reference cases). The general principle of CBR is to treat a new problem (target case) by remembering similar past experiences (base case). This type of reasoning rests on the assumption that if a past experience and new circumstances are sufficiently similar, then everything can be explained or applied to past experience (base case) and remains valid when applied to the new situation which represents the new problem to solve. From a very global view, the CBR uses a basis of experience or case, a mechanism for searching and retrieving similar cases and a adaptation mechanism and evaluation solutions of selected cases emanating in order to solve the specified problem (Figure 2).

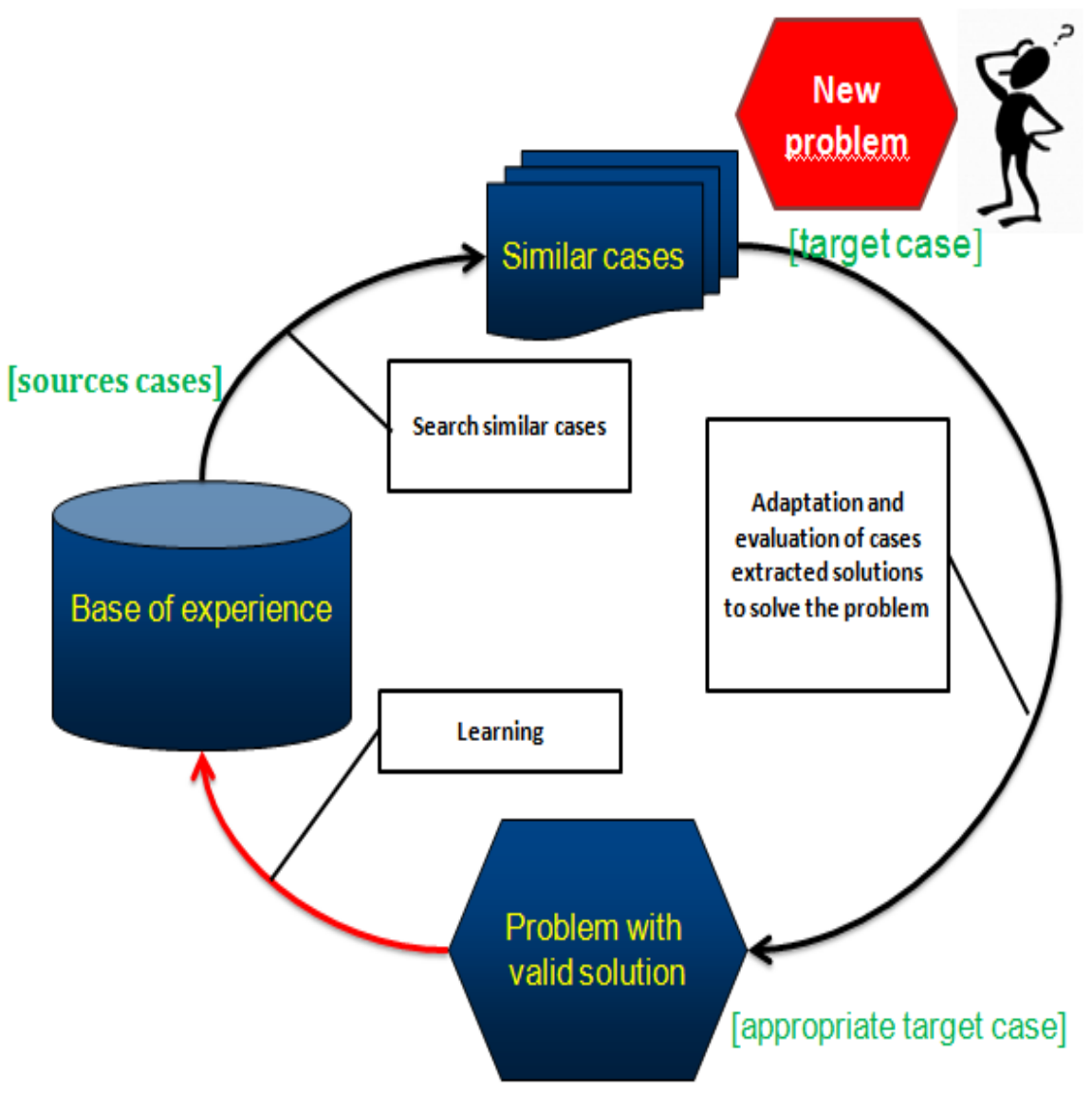

Fig. 2 cycle case based reasoning (CBR) [8]

Generally, the reasoning cycle consists of seven main steps, detailed below:

1) The development of an index, some indexes are assigned to the "new problem (target case) to characterize it. These indexes are formed from the information we extract from the new problem. They are used to find similar cases.

2) The case-finding (or remembering): Indexes and matching rules are used to try to find a previous case (source case) similar to the current problem (target case).

3) Adaptation: the old solution is modified to be adapted to the new situation. The result is a solution to the new problem again.

4) The test: the solution is tested to determine its suitability to the problem.

5) Memorization: If the proposed solution is successful, it incorporates the new case (with the features of the proposed solution) in memory of the event. It is this stage of learning that can enrich the knowledge of the system.

6) The explanation: if the solution fails, we try to explain it by looking for causes of failure.

7) Correction: Knowing the cause or causes of failure, the solution is corrected before repeating the test.

The work of [Slade 91], [Harmon 91], [Rougegrez 93], [Kolodner 89, 91, 92, 93], [Beauboucher 93], [Mott 93] [Pinson 93], [Smail 93] and [Lieber 93] trace a fairly complete development of research in the field of case based reasonings.

Started ,we have demonstrated that this particular process Rex suffers from lack of adequate and relevant tools to exploit all the knowledge obtained. Our study, is in this context, and aims to develop a model of feasibility based on case based reasoning to tackle this problem. The developed tool is presented below.

## IV. FEASIBILITY OF MODEL BASED ON THE ARGUMENT FROM CASE

### *A. THE PROPOSED APPROACH:*

The approach taken to develop this tool involves two major activities:

- To demonstrate the feasibility and merits of our approach, we used a database that includes 70 accident scenarii work of acquiring knowledge INRETS (related to the problem of collision) [14]

- To exploit the knowledge stored in having recourse to case based reasoning in order to identify and deduct know-how that can help stimulate domain experts to propose solutions or preventive measures and / or protection

Before presenting the general architecture of the developed model, we should define what is meant by an accident scenario.

### B. *CHARACTERIZATION OF AN ACCIDENT SCENARIO*

An accident scenario describes a combination of circumstances that can lead to a undesirable or even dangerous situation. It is characterized by a context and a set of events and settings [15,16]. The acquisition of knowledge has led to the shaping of a particular model based on the identification of eight parameters describing an accident scenario (Figure 3).

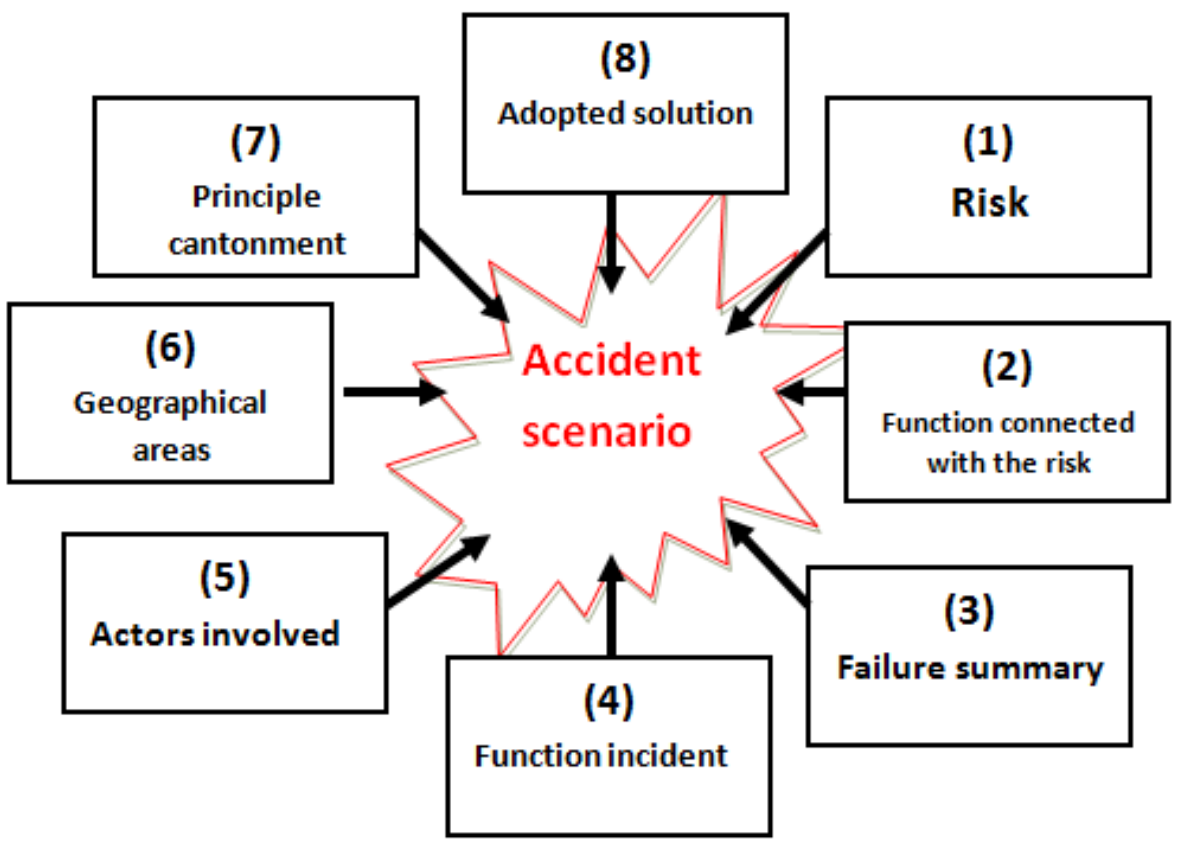

Fig.3: Characteristic parameters of an accident scenario [15,16]

### C. *FUNCTIONAL ARCHITECTURE OF THE MODEL OF THE DEVELOPED TOOL*

Very briefly, the developed model is built on the 3 following modules:

• A database of accident scenarii representing historical knowledge from Rex.
• A man-machine interface for archiving, editing and entering new scenarii for evaluation.
• A module that represents the process of case based reasoning.

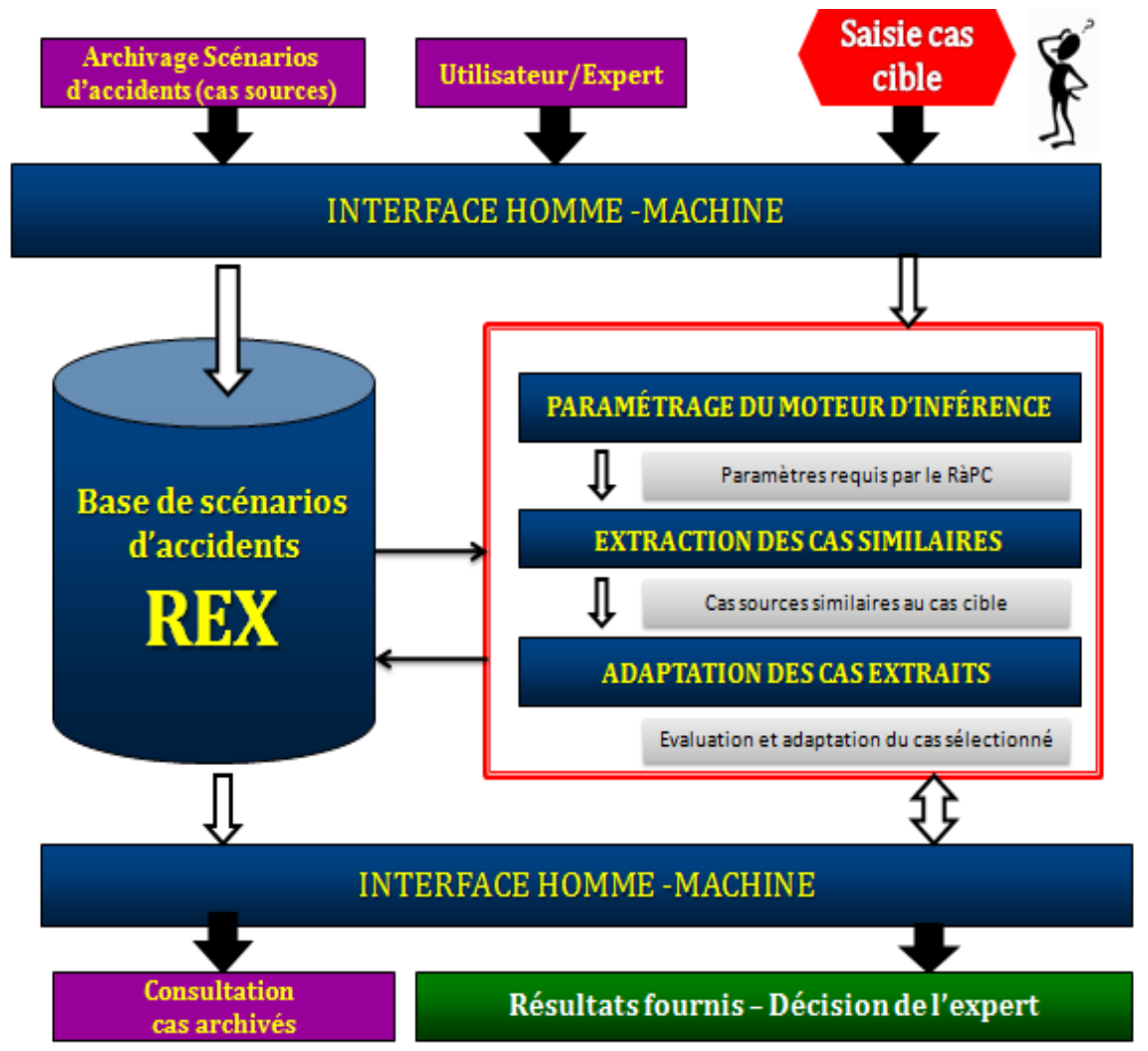

Fig. 4 Functional architecture of the developed model

In order to show the feasibility of the developed tool, it seems wise to use a sample application that shows the progress of an example of use.

*D. EXAMPLES OF APPLICATION (USAGE):*

Using the developed model requires the passage through the 4 phases

-Entering a new target case
-Extraction of the most similar cases
-Adaptation and evaluation of scenarii extracts
- Learning and enriching of the database

*a. SEIZURE OF A NEW TARGET EVENT*

Once we have built all accident scenarii which make our base of experiences, we must now begin the cycle of case based reasoning (CBR) by entering a new target case (unsolved problems) the interface Figure 5.

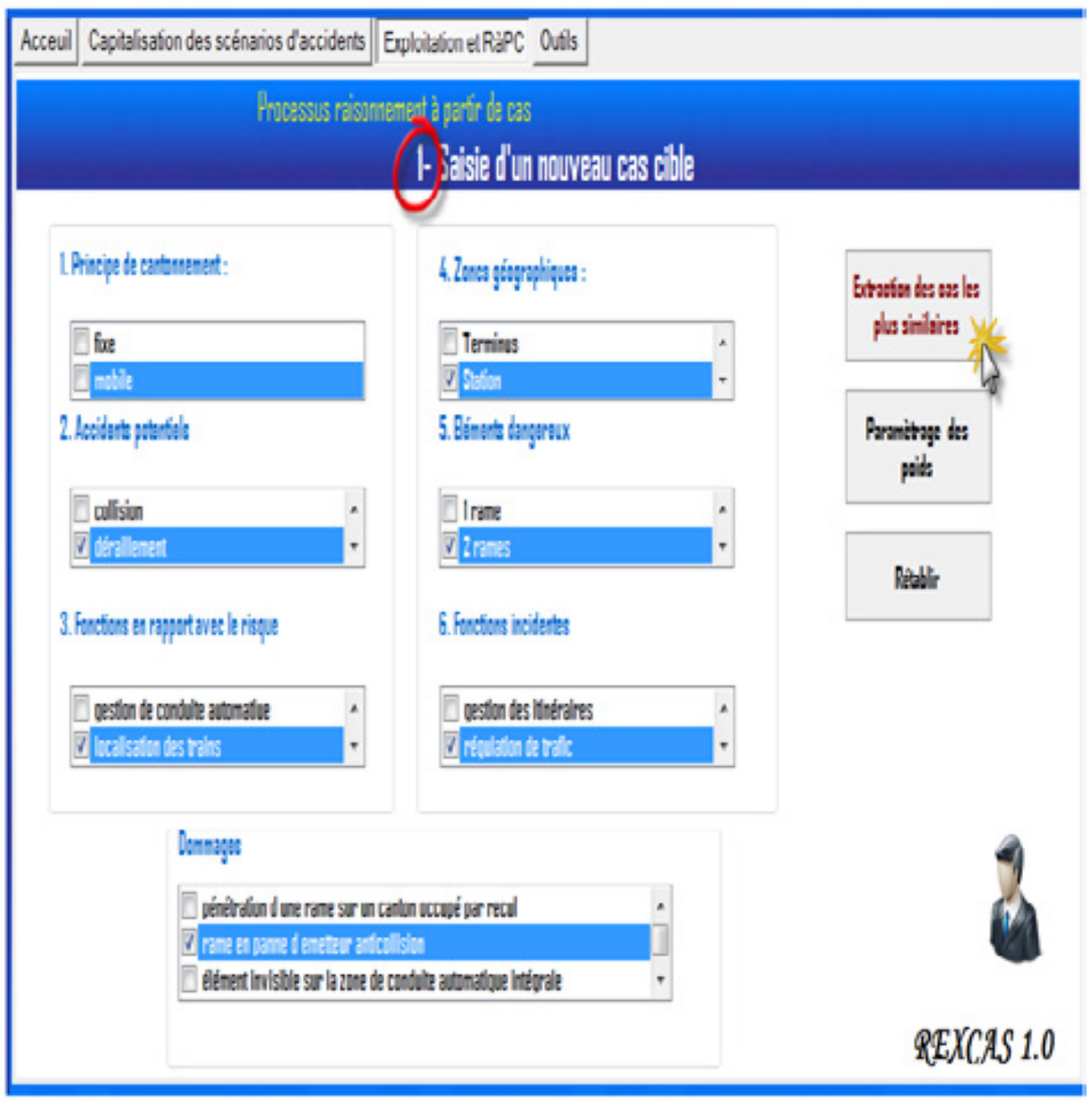

Fig.5 Entering a new target case

*b. EXTRACTION OF THE MOST SIMILAR CASES:*

Before you start your search for the most similar case from the case base (base of experience), you can adjust the search by editing the weight of each attribute of a script. The tool proposes a similarity measure as the compensatory model [17]. This similarity measure is called "overall similarity". The similarity of case is a percentage calculated from the weight of each descriptor and all similarity measures performed descriptor by descriptor. Thus, if a case has 3 descriptors of which 2 are 100% similar to the C.

with the target case and the third not at all (0%), then C will be similar with the target case to 66% if all the descriptors are the same weight : ((100 × weight descriptor 1 + 100 × weight descriptor 2 + 0 × weight descriptor 3) / 3 = 66). The user can intervene in several ways to calculate the similarity between two attributes. It may specify the descriptors that should not be taken into account when calculating. It can give a weight vector to indicate the relative importance of a descriptor against other by clicking the button configuration known weights Figure 5. In our example, we chose to extract only the 5 most similar cases; below Figure 6

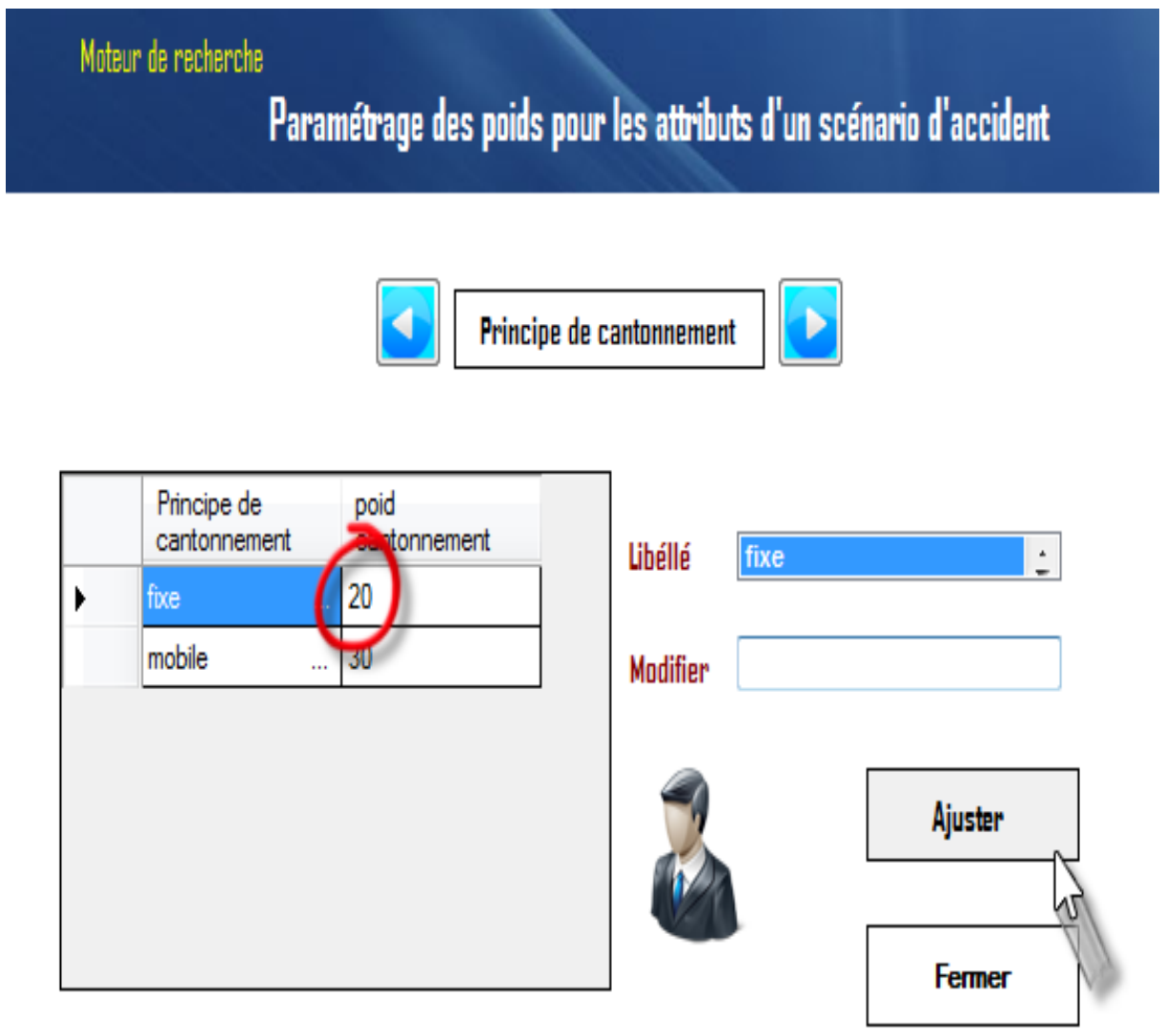

Fig. 6: Setting weights for the descriptors of an a scenario accident

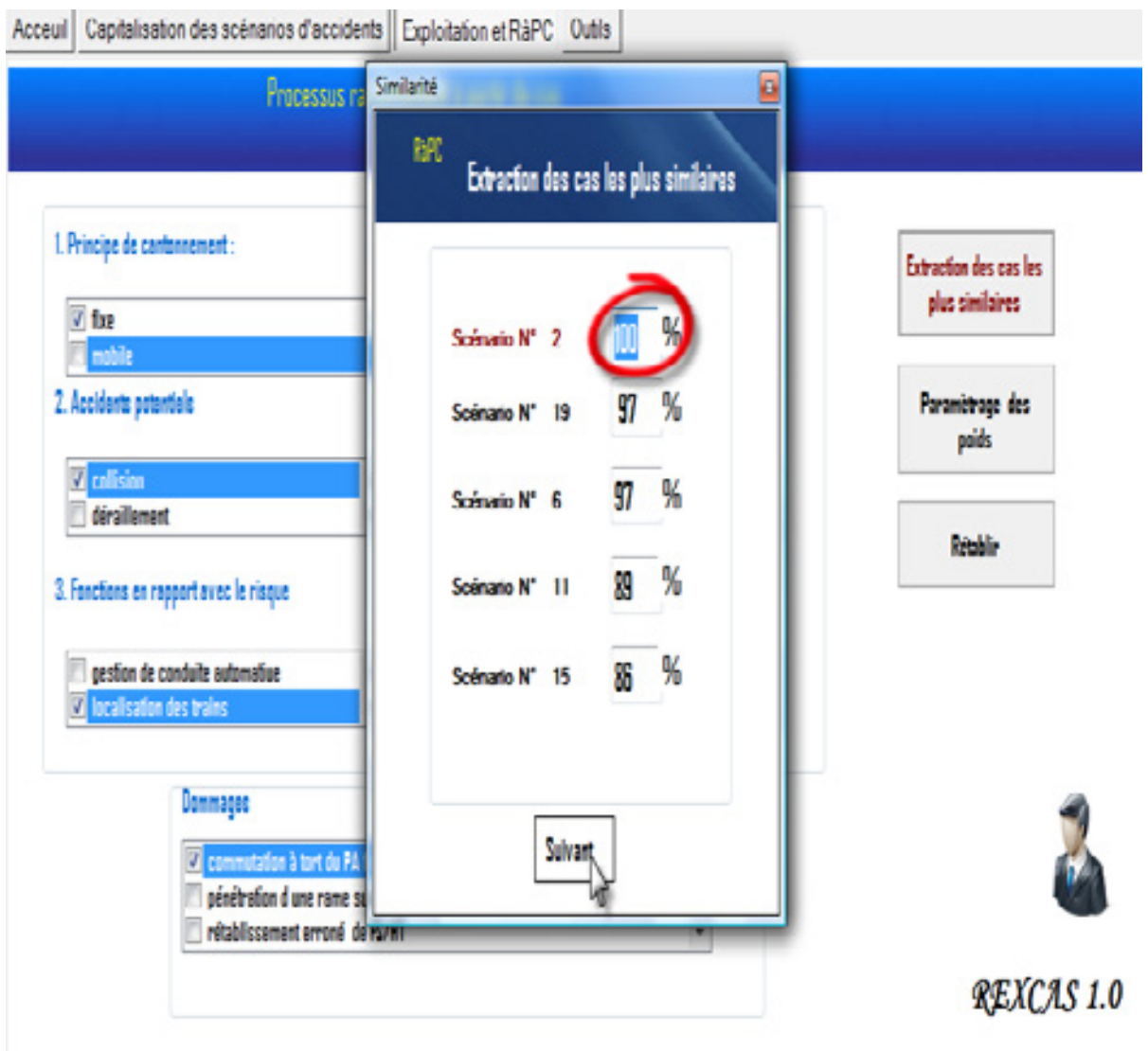

Fig. 7: Extraction of the five most similar cases

This step allows to find the historical accident scenario (event sources) closest to the scenario to the considered (target case). The screen shown in Figure 7 shows, for our example, the results of the research for similar cases.

The target case is recalled in the right column while the left column offers the first 5 cases and more like the middle column can show a similar case (here the case 2).

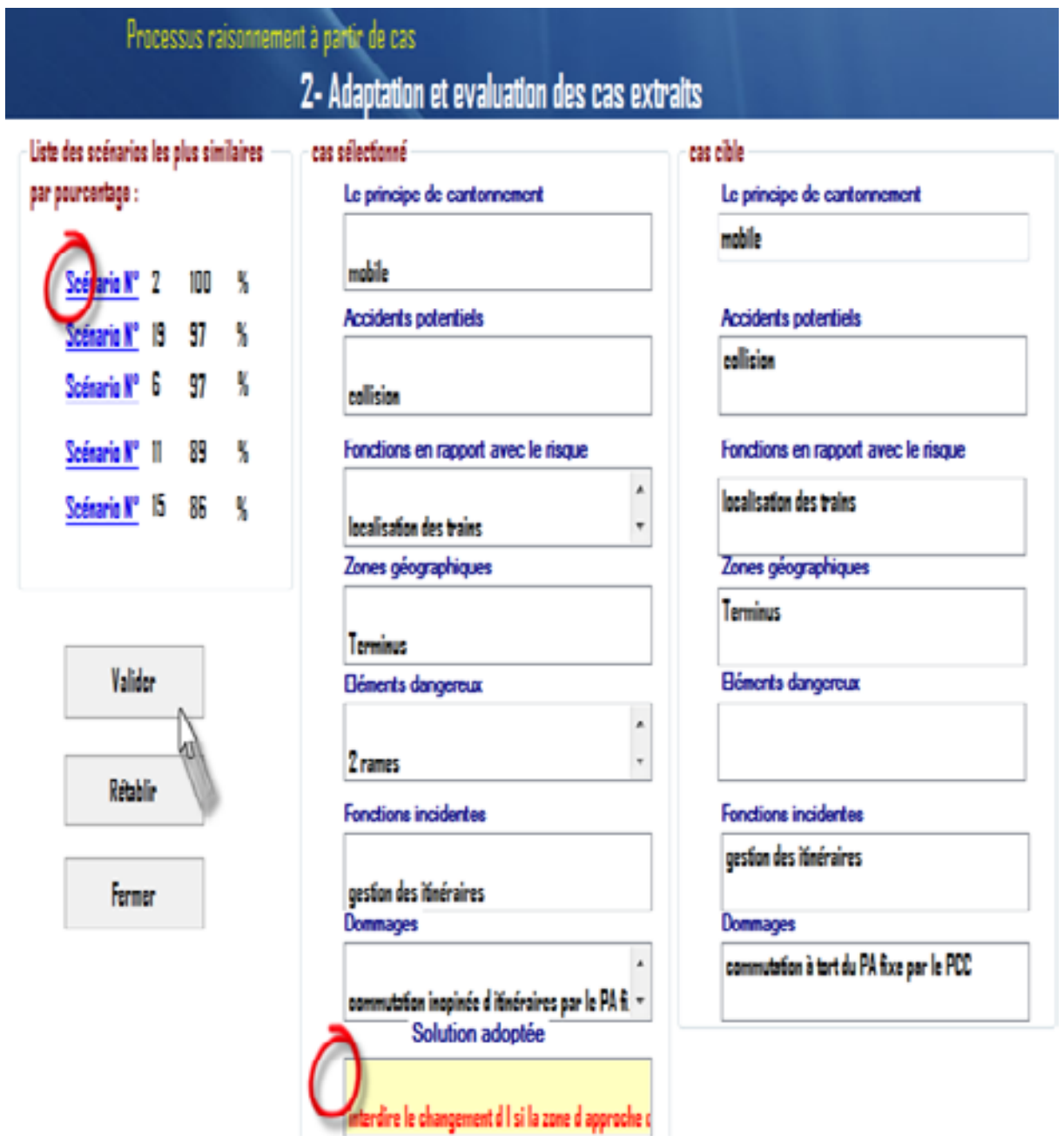

Fig. 8: Selection of cases retrieved for adaptation

## c. *ADAPTATION AND EVALUATION OF SCENARII EXTRACTS:*

Our tool does not propose coping strategies, this is left to the user. With the screen shown in Figure 9. The user can see the value taken by attribute concept "solution adopted in each case similar and an select his self to provide value to the attribute" concept "for the target case . The user can also use the technique of the vote. In our example, the tool proposes a single value for the attribute 'solution adopted ": Check the actual docking. Thus, the domain expert can adjust if the "Check the actual docking" as a solution to the problem or he can choose a solution from the base up and in extreme cases he may propose a new solution to solve the problem and simultaneously enrich the basic solutions.

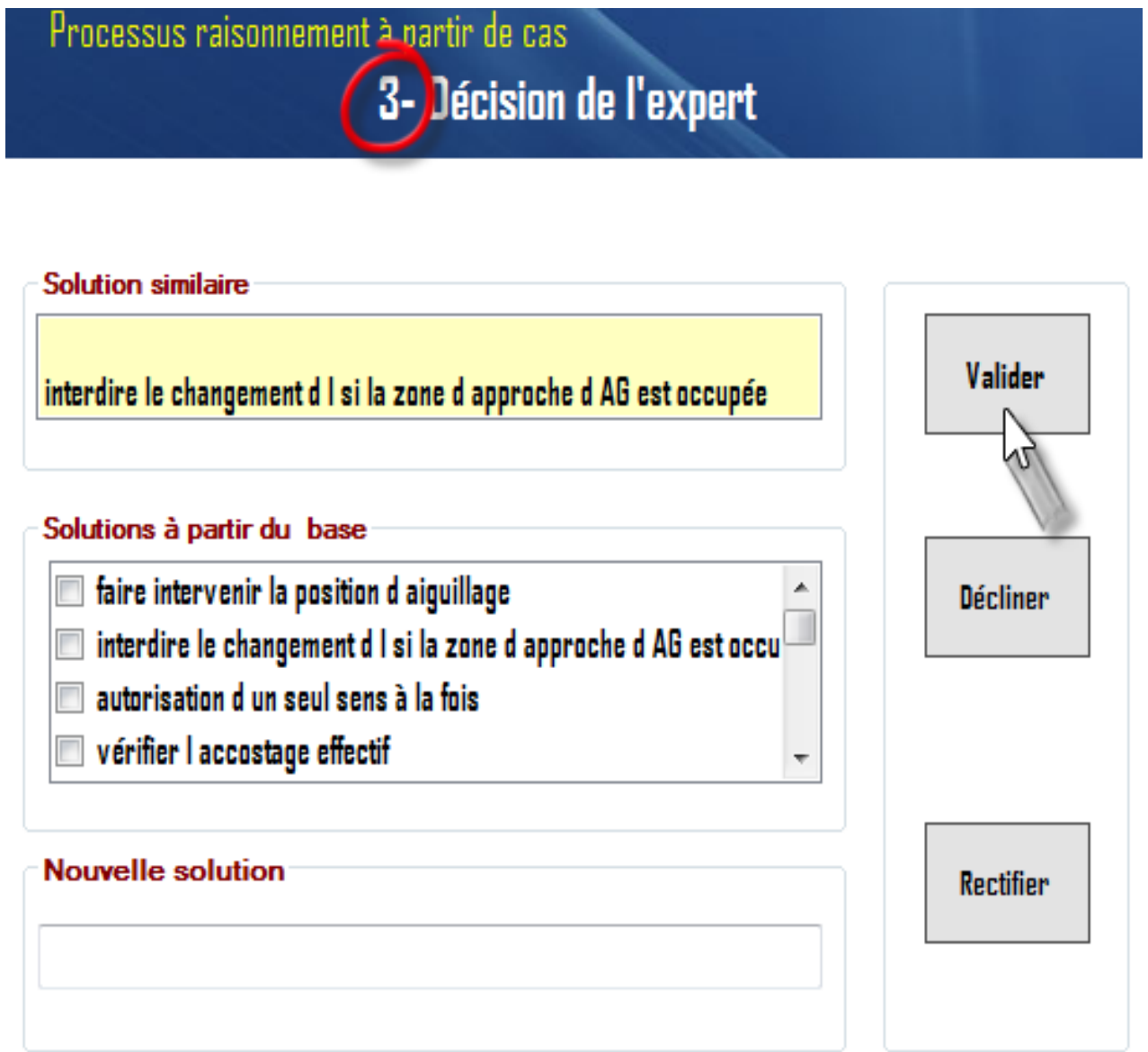

Figure 9: Adaptation and decision of the expert

### *d. LEARNING AND ENRICHMENT OF THE BASE:*

This final step in upgrading is to make learning by adding the missing attributes like number, title and class of cases suitable target in the base case (scenario historical accidents). In this learning tool is incremental because the new case will be integrated into the case database sources for future use. see Figure 10.

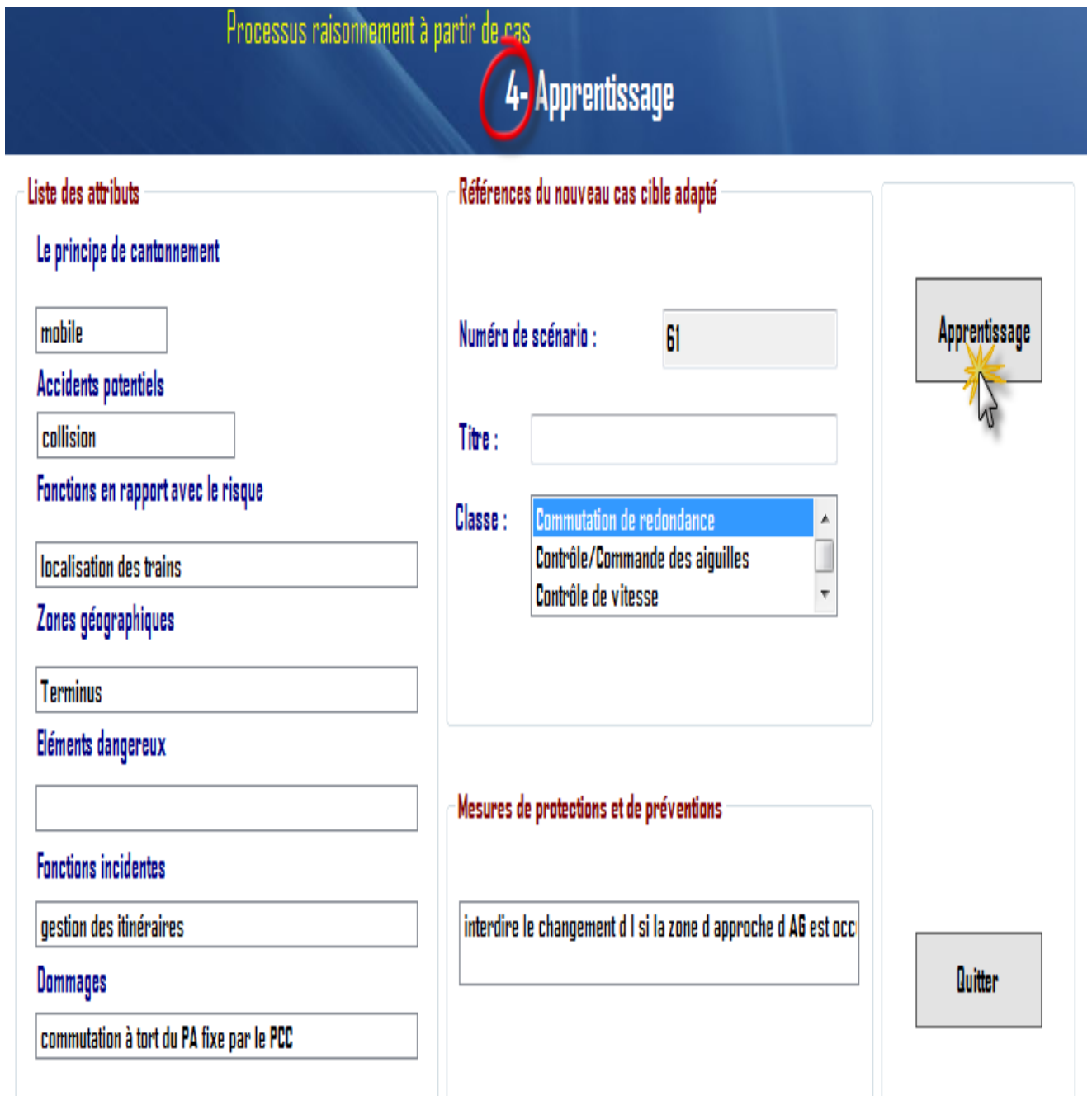

Fig.10: Learning and enrichment of the database

## V. CONCLUSION AND PERSPECTIVES

This paper is to show the contribution of case based reasoning (CBR) to exploit the experience feedback. This approach has an undeniable and original interest because this is , in our opinion , the first research on exploitation of knowledge in this field of safety. In addition the interest of this work lies not only in terms of the acquisition, representation of accident scenarii but also at the level of knowledge exploitation of accident scenario in order to help and assist experts in their crucial task of analysis and evaluation studies of system security.

However, this tool requires some improvements and extensions. These improvements include the choice of evaluation criteria and the accuracy of results; namely , the treatment of missing values, improved adopted inference strategies and improvement and validation of formalism representation of accident scenarii . Indeed, this model can be improved and is still only a working basis for the definition of a generic model acceptable to all actors involved in the development of guided transport systems. I would like to thank all members of the ACT7 group of Tunisian researchers in the Transport Safety chaired by Professor HADJ MABROUK .